# Architecting and Visualizing
# Deep Reinforcement Learning Models

Alexander Neuwirth and Dr. Derek Riley
Department of Electrical Engineering and Computer Science
Milwaukee School of Engineering
1025 N Broadway, Milwaukee, WI 53202
neuwirtha@msoe.edu, riley@msoe.edu

## Abstract

To meet the growing interest in Deep Reinforcement Learning (DRL), we sought to construct a DRL-driven Atari Pong agent and accompanying visualization tool. Existing approaches do not support the flexibility required to create an interactive exhibit with easily-configurable physics and a human-controlled player. Therefore, we constructed a new Pong game environment, discovered and addressed a number of unique data deficiencies that arise when applying DRL to a new environment, architected and tuned a policy gradient based DRL model, developed a real-time network visualization, and combined these elements into an interactive display to help build intuition and awareness of the mechanics of DRL inference.

# 1 Introduction

In recent years, deep reinforcement learning has been used to address a wide variety of impactful problems across numerous fields. To meet the growing interest in educational visualization related to this emerging technology, we sought to build a DRL model to drive an AI agent for Atari Pong as an accessible and visually interesting demonstration of how DRL can be applied. Additionally, we set out to develop a visualization that communicates the inference of this model to help build intuition on high level DRL concepts and better understand and tune the training of our model.

Deep reinforcement learning was first popularized through "Playing Atari with Deep Reinforcement Learning" (Mnih, et al. [1]) in which a single model and training setup was used to train agents with superhuman performance in three different Atari games. The researchers utilized deep Q-learning to optimize a policy based on the estimated reward of actions given a specific state. Since then, several works have expanded upon these ideas.

Andrej Karpathy's "Pong from Pixels" blog post [2] extended [1] with a simplified approach, utilizing a REINFORCE algorithm to beat Atari Pong using a two-layer dense network with a single hidden layer of 200 neurons. This technique forms the basis of the training approach used in our work, and seeks to directly optimize the action policy by applying small adjustments to the policy each episode.

The visualization component was inspired in part by the visualization of the "MarI/O" agent, created by video game commentator and live streamer SethBling [3]. This agent uses neuroevolution of augmenting topologies to genetically construct neural networks based on fitness of agents.

While a number of papers, tutorials, and online examples exist that apply DRL to a Pong environment, project requirements necessitated flexibility of elements that existing implementations cannot support, including full control of one or both paddles by the AI agent or human players and the use of a physics engine that can be reparameterized between or during games. An innovative data pipeline and training simulation strategy were developed in this work to enable improved performance of training, better flexibility for model variations, and ease of future tuning to this model.

In this work we present an end-to-end DRL application that produces useful visualizations of deep reinforcement learning training and inference, including the engine to simulate the Pong environment, the DRL model, and a policy gradient based training implementation.

## 2 Project Structure

In order to effectively train a DRL agent and visualize the completed model in a way that provides useful insights, we created three modular components: (1) a custom Pong environment to track and increment game state (see section 2.1), (2) a DRL agent that accepts a downsampled and preprocessed image of the game as input and produces a probability distribution across the action space (see section 2.2), and (3) a real-time visualization tool that accepts the model weights and biases from the DRL agent and each frame from the pong environment, rendering a display of the model's inference and most significant activations given each frame (see section 2.3). Figure 1 provides a high-level overview of the components and their relationships.

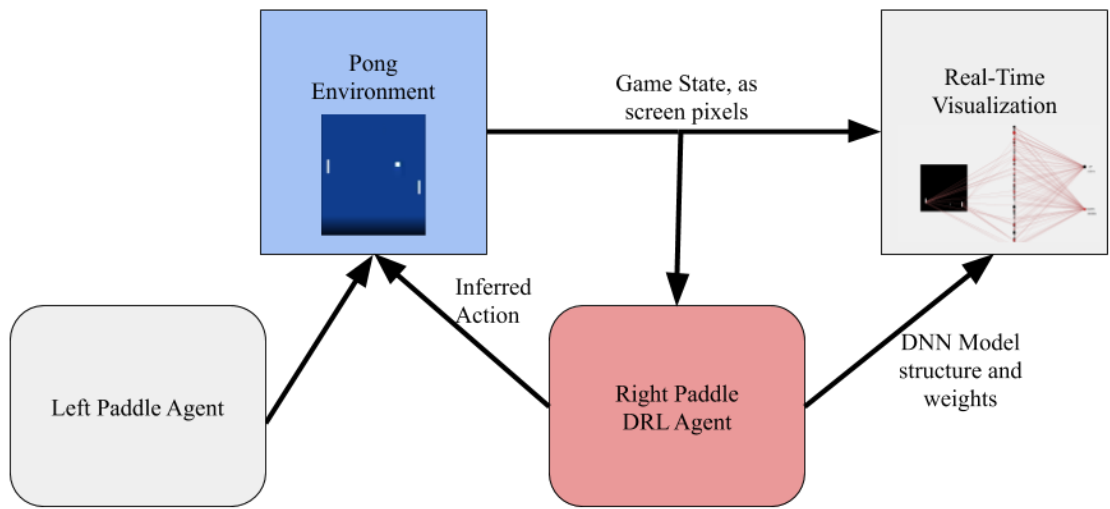

Figure 1: Relationships between environment components

### 2.1 Custom Pong Engine

Pong is a video game originally created by Atari [4] and inspired by table tennis (see figure 2). In the game, two players each control a paddle that moves up and down in front of a goal. A ball bounces across the screen, and each player attempts to collide their paddle into the ball to bounce it into the opponents' goal. The game's simplicity makes it a fairly common environment for reinforcement learning research.

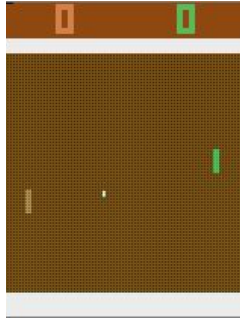

Figure 2: The OpenAI Gym Atari Pong Environment

A pong engine is necessary to maintain a game state (ball and paddle positions and velocities). At each game timestep, it accepts an action (move the paddle "up", "down", or "none") to progress to the next timestep, and then emits the state image and score value (1 for a score, -1 for an opponent score, and 0 for neither). The pong engine commonly used in RL research is an emulated Atari version of Pong included within OpenAI Gym [5], a Python library for RL research that gives a simple and common API to several different environments.

We created a new Pong environment (see figure 3) that differs from the OpenAI Gym Atari emulator in three major ways:

1. Our engine is able to accept actions and output rewards for one or both paddles simultaneously. This allows the model to compete against a human player or against another model.

2. Our custom environment has finer timestep resolution than the Atari Emulator. Using paddle movement distances for comparison, each timestep in the OpenAI Gym Atari environment captures the same amount of game time as approximately ten timesteps in our environment. Additionally, the OpenAI Gym environment processes a fixed 2-4 frames at random between each timestep, but our environment exposes this as an additional hyperparameter.

3. Each aspect of our environment's physics and game mechanics can be easily configured, including ball and paddle sizes and speeds, field size and dimensions, and initial game state.

We found it necessary to create a new environment because the interactive visualization envisioned requires that one paddle can be controlled by the player while the other is controlled by the DRL agent, and potentially would involve changing physics constants or game state directly. Since the API provided by the OpenAI Gym Atari environment is very limited, only supporting the exact game configuration and process provided by the Atari emulator and only accepting actions for one paddle at each game timestep, the only way to support the flexibility we desired was to create our own environment.

Our environment and game logic are written in Python, utilizing NumPy to draw output frames by setting pixel color values over a region of a 3D height-width-color array with array slicing. When running without artificial time delays to train a neural network agent, our environment completes a full episode (played until one side reaches 21 points) in two to four seconds, which was experimentally determined to be about seven times faster than the Atari emulator implemented by OpenAI gym..

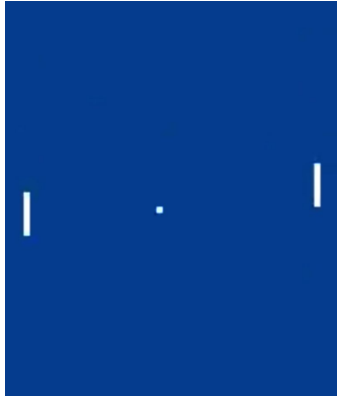

Figure 3: Our custom pong environment

## 2.2 Deep learning framework

The deep learning framework used for our pong agent was written using Keras [6] and based on the design demonstrated by Andrej Karpathy in his "Pong from Pixels" blog post [2]. The model's input is the preprocessed pixel state of the game screen: downsampled by a factor of 2, cropped to include only the rectangle of valid paddle and ball locations, and filtered so the ball and paddle are represented as "1" and all other game features are "0". To capture paddle and ball velocity, the actual model input is the pixel-by-pixel difference (see figure 4) between each frame and its previous frame. The input at each timestep is therefore a matrix sized at half the dimensions of the game area (80x80 for Atari environment, 80x96 for our environment) where each pixel $p \in \{-1, 0, 1\}$.

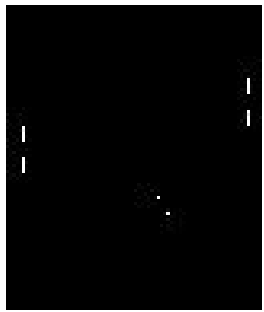

Figure 4: An example model input, taken as the difference between two frames

In most of the experiments discussed in this paper the model outputs a vector in two dimensions, representing "up" and "down", although some experiments included a third "none" dimension. Categorical cross-entropy loss was used to output a probability distribution over each action. Binary cross-entropy would have been suitable for the experiments using only two actions, categorical cross-entropy was used for all trials so that actions could be freely added to the output layer without changing the loss function. Between the input pixels and output layer, 1-4 hidden dense layers of 50-300 neurons were used depending on the experiment. These numbers were selected based on the 200-neuron hidden layer in Karpathy's Pong agent [2] and hand-tuned to optimize performance.

The model was trained on batches of between 1 and 25 episodes, configured before each experiment. Each episode consists of a pong game played until one side reaches 21 points.

The DRL training process uses a REINFORCE algorithm [7] to directly optimize the policy. The model was assigned randomly initialized weights. A batch of games is played with the policy set by this model, and each screen state, action, probability vector, and reward is stored. At the completion of the batch, the experiences are processed as follows:

1. *Rewards are discounted.* As rewards are only allocated as "1" for timesteps where the agent scored, "-1" if the opponent scored, and "0" otherwise, rewards are discounted backwards to assign blame to actions leading up to rewards. This is done by iterating backwards through the game memory and recursively reassigning each reward value of zero to the value of the next timestep's reward multiplied by a discount factor $\gamma$. Across all experiments, $\gamma$ was held constant at 0.99.

2. *The gradient is calculated for each timestep.* To calculate the gradient, the probability output is subtracted from a one hot encoded vector of the selected action for each timestep, then multiplied by the discounted reward for that timestep.

3. *Labels are assigned to each state.* Each state is labeled with the originally assigned label, plus the gradient calculated in step 2 times a learning rate $\alpha$.

The model is then trained on the labeled batch for one epoch. The updated weights are used to play another batch of episodes and this process is repeated until loss no longer improves.

## 2.3 Visualization

We constructed our visualization with the intent to clearly show how DRL agents perceive and act on their environment and build intuition on how the input is processed in the hidden layers. The visualization renders an image for each game frame and draws each individual neuron in the model as a circle. Weights are visualized as lines spanning from neuron to neuron or from neuron to pixel of the input image, and every numerical weight is hidden by default. To reduce visual noise, all weights under a threshold are set to never render at all. During each episode, neurons that activate over a certain threshold light up in red, and all of the weights extending from active neurons and nonzero pixels also render in red. Each output neuron changes to an intensity of red corresponding to the strength of its activation, and confidence percentages and labels are given by each output option (see figure 5.)

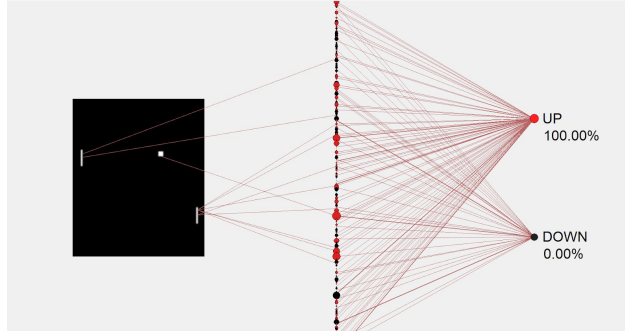

Figure 5: The visualization UI

This visualization was written in Python using the tkinter canvas [8], and integrates with the RL agent and simulation environment. Once initialized with the agent's model structure and weights, the visualization renders the model's interaction with any game state passed to it.

## 3  Experiments

We initially trained our DRL agent to run against the custom emulator with a single hidden layer of 200 neurons, utilizing a batch size of 10 episodes, a learning rate of 0.001, and a ball diameter of 2 pixels. All experiments use this structure as a baseline.

### 3.1  Opponent Strategy

For all training, our environment's left paddle was controlled by a hand-engineered AI with full knowledge of positional game state. This AI was first programmed with the following policy:

1. If the ball is moving left (toward the AI), then track the ball - moving up if the ball is above the paddle, and otherwise moving down.

2. Whenever the ball is traveling to the right (away from the AI), randomly choose to move up or down.

For ease of reference, we refer to this strategy as "partially-tracking". Competing against the partially-tracking opponent proved to be a surprisingly difficult task, and the baseline agent was unable to score above 15 points after 10,000 training iterations. For comparison, we modified our hand-engineered opponent to use a "consistently-tracking" strategy - tracking the ball at all times, whether it is moving left or right. (This behavior is consistent with the OpenAI Gym Atari environment's computer opponent.) Agents trained against a consistently-tracking opponent appeared to find a significantly better policy than those trained against the partially-tracking opponent, frequently scoring 21 points to win episodes after about 6,000 training iterations. However, once the trained model was set to compete against a partially-tracking or human opponent, it performed no better than randomly selected actions. This seems to suggest that agents trained against a consistently-tracking opponent are prone to generalize poorly, using the opponent's paddle position as an additional indicator of ball position. For this reason, agents trained against a consistently-tracking opponent are not suitable for use in an interactive visualization that competes against a human opponent. This problem is addressed in section 3.4.

In order to tune model structure and hyperparameters in parallel with investigating this problem. All experiments were run against the consistently-tracking opponent unless otherwise specified.

## 3.2 Learning Rate

In an effort to speed training time, we trained an agent with a learning rate of 0.01, ten times higher than the baseline. This agent reached minimum loss in about 100 batches and achieved a maximum score of 15 after about 4,000 batches, as opposed to the baseline which reached both minimum loss and maximum score simultaneously after around 6,000 batches. The baseline learning rate also achieved significantly better performance, frequently winning episodes by scoring 21 points. These results seem to suggest that a dynamic learning rate that scales down over time could improve convergence.

## 3.3 Network Size

Several different model architectures were tested with the goal of improving on the baseline. Only relatively small and shallow dense networks were tested, in order to allow the final agent to be easily visualized completely.

Several different models' scores are depicted in figure 6. Each plot shows trailing average accuracy of 100 episodes to filter out noise. Note that the 200-200-100 structure significantly outperforms the others.

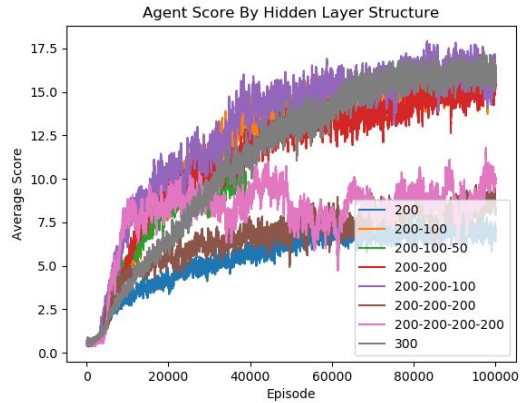

Figure 6: Performance of different network architectures. Labels indicate the number of neurons per hidden layer, with layers delimited by hyphen.

These results seem to indicate that deeper networks with larger initial dense layers narrowing into smaller layers are most suited to discovering a good policy in this environment.

## 3.4 Ball Diameter

As discussed in section 3.1, agents trained against an opponent that does not reliably track the ball's movement with the paddle (such as our "partially-tracking" AI) were unable to discover a winning policy. With the hypothesis that increasing the ball's visibility could serve a similar function to a consistently-tracking paddle as an added state indicator, we fit the baseline model against an environment where the ball's diameter is 6 pixels instead of 2 pixels, with all other factors identical to the baseline.

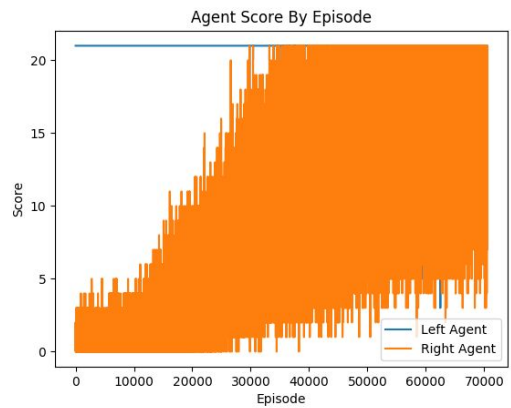

Figure 7: Training scores of the baseline agent with 8px ball diameter

This proved very effective: the agent trained in this environment was able to outperform all other models discussed in this section (see figure 7) and was even able to generalize well enough to compete against a human opponent.

# 4 Challenges

A wide variety of interesting data deficiencies were encountered while fitting the DRL model to our environment. Several of the more interesting challenges and the considerations taken to address them are enumerated here.

## 4.1 Loss of information with frequent frame sampling

As the Pong environment's frame dimensions are quite small and each render is only a discrete approximation of the ball's movement, it is possible for a slow moving ball moving at specific angles to be rendered at the exact same pixel locations for two consecutive frames if render samples are taken frequently enough. Since the model's input is the pixel difference between frames, this scenario causes the ball's position to be subtracted from itself, rendering the ball invisible to the model (see figure 4.) This data deficiency severely harmed performance in early models, and was addressed in our approach by skipping several frames between model timesteps to ensure the ball moves enough to change render location.

## 4.2 Inability to detect stationary paddles

As another consequence of using frame differences as model input to track movement, stationary paddles do not appear in the model input. This caused poor performance when either the DRL agent or the hand engineered opponent chose to remain stationary. We considered a number of methods of addressing this deficiency, including only subtracting the frame columns between the paddles or including the information in some other way, but ended up simply removing the "None" action from both models for simplicity.

## 4.3 Downsampling data loss

As the screen state is downsampled by a factor of two as part of preprocessing and the ball is fairly small relative to the rest of the screen, downsampling originally caused the ball to blink in and out of existence while moving along certain trajectories. This was eventually resolved by increasing the ball size to the point where this no longer occurred. (This is unrelated to the experiments regarding increased ball size discussed in section 3.4, which experimented with ball sizes even larger than the dimensions that prevented downsample loss.)

## 4.4 Overfitting to specific opponent AI

Even when the DRL agent could successfully compete against a hand-engineered opponent, it often performed far worse against a different hand-engineered opponent or human opponent as new states were encountered that never appeared during training. In an effort to address this, a new environment was created in which the agent is tasked to play against an empty goal while balls spawn with random positions and velocities from the other side of the screen. This problem and related experiments are discussed in greater detail in section 3.1.

# 5 Conclusion

We present the following contributions:
· A new Pong environment with a much higher degree of configurability than the current standard, including the ability to compete against a human opponent.
· A visualization tool capable of delivering interesting insights on the behavior of small fully-connected neural networks in real time.
· A dense model with only 200 hidden neurons capable of generalizing well enough to compete against both human or AI opponents.
· The insights necessary to further refine our model and inform further research.